\documentclass[english]{maciarticle}
\usepackage{amsmath,amsbsy,amscd,amssymb,graphicx,epsfig,color}
\usepackage{multirow}
\usepackage{bbm}
\usepackage[utf8]{inputenc}
\usepackage[english]{babel}
\usepackage{yhmath}
\usepackage[center]{subfigure}
\usepackage{multirow}
\usepackage{graphicx}
\usepackage{subfigure}

\usepackage{authblk}

\begin{document}

%\maketitle
\thispagestyle{empty}
%\newpage

% display page numbers in the headings. Start with roman numerals %
\pagestyle{headings}
\setcounter{page}{1}
\title{Pattern recognition in SAR images using
Fractional Random Fields and its
possible application to the problem of
the detection of Oil spills in open sea
}
\date{}
\author[1]{ Agust\'{i}n Mailing} 
\author[1]{ Segundo A. Molina} 
\author[1]{ Jos\'{e} L. Hamkalo}
\author[3]{ Fernando R. Dobarro}
\author[1,2]{ Juan M. Medina}
\author[1,2]{ Bruno~Cernuschi-Fr\'{i}as} 
\author[3]{ Daniel A. Fern\'{a}ndez}
\author[3]{ \'{E}rica Schlaps}
\affil[1]{ Facultad de Ingenier\'{i}a, Universidad de Buenos Aires, Buenos Aires, Argentina,
\textcolor{blue}{jmedina@fi.uba.ar}}
\affil[2]{ Instituto Argentino de Matem\'atica "A. P. Calder\'on", IAM, CONICET, Buenos Aires, Argentina.}
\affil[3]{ Universidad Nacional de Tierra del Fuego, Ant\'artida e Islas del Atl\'antico Sur, Instituto de Ciencias Polares, Ambiente y Recursos Naturales, Ushuaia, Argentina,
\textcolor{blue}{fdobarro@untdf.edu.ar}}

\maketitle

\begin{abstract} In this note we deal with the detection of oil spills in open sea via self similar, long range dependence random fields and wavelet filters. We show some preliminary experimental results of our technique with Sentinel 1 SAR images.

%\bigskip \bigskip\bigskip \bigskip 

$ $

\end{abstract}


\textbf{AMS class:}
60G22, 60G35 (Primary), 42C40, 60G20 (Secondary)

%\bigskip \bigskip

%\section{Introduction}
Assuming that an intensity SAR image is modeled by a self
similar random process \cite{Cohen} we propose a possible rather simple analysis scheme for the detection of oils spills in open sea. Moreover as this is essentially a smoothness or complexity analysis technique of the image it can be potentially applied to the recognition and classification of other patterns.
To explode notions related to self similarity and fractals
related to SAR image processing (and optical images) is not
completely new. In e.g. \cite{MarghanyHashim2007,MarghanyHashim2011} computing an approximate
Box/Hausdorff fractal dimension of the neighborhood of each
pixel some possible algorithms for oil spill detection are
proposed. Indeed, since "fractal" is a more or less vague
term, several underlying models with varying results and
computational complexity can be proposed. As suggested in
\cite{MarghanyHashim2007,MarghanyHashim2011} box counting techniques have a rather good
performance in solving this detection problem. However this
may be not the case from the the point of view of the
required computational time to perform this task. In
contrast, Fourier analysis methods or similar as wavelet
analysis \cite{Cohen} are potentially much simpler and fast since
there exist several optimal FFT algorithms. Here, among all
possible models of self similar random fields to model the
image we propose the class of the stationary (or locally
stationary) $\dfrac{1}{f}$-processes \cite{Cohen}. These fields were introduced
by A. Kolmogrov in a first statistical study of turbulent flows.
An $X=\lbrace X(p),p\rbrace$ random field of these class, with $p$ in $\mathbb{R}^{2}$,
has an spectral density of the form (or asymptotically near
the frequency $\omega =0$) \cite{Rosenblatt1985}
\begin{equation}\label{eq:1}
S_X(\omega)= \dfrac{C}{\Vert \omega \Vert ^{a}}.
\end{equation}
The exponent $a$ gives a quantitative measure of the the
correlation of the random field. For greater $a$, $X$ is more
correlated and in particular for appropriate values of $a$, $X$ displays \emph{Long Range Dependence}. In contrast, for $a \longrightarrow 0$, $X$ is nearly white noise.
The parameter $a$ is linearly related to the so called Hurst
parameter \cite{Cohen}, which measures the self similarity of the random field. Finally, at least in the Gaussian case, 
$a$ is also related to the Hausdorff dimension of the graph of $X$ and the roughness of it.

In this context, we shall assume that the portion of image
containing the oil spill fluid will be approximately behaving
as a $\dfrac{1}{f}$-random field with a certain parameter $a$. Waves are
supposed to be attenuated in contact in the presence of oil,
which under suitable conditions, will traduce in a smoother
and highly correlated surface in comparison to non
contaminated open sea water. The proposed detection
scheme is basically to test if the possible oil spill area follows
model \eqref{eq:1} for some $a$. To do this, given any finite energy
filter $h_{1}(p)$ with a Fourier transform $H_{1}(\omega)$ we consider a
scaled version of it $h_{2}(p):=h_{1}\left(\dfrac{p}{2}\right)$ and if $X$ is our
image/field we obtain two filtered (convoluted) versions of it,
namely:
\begin{equation*}
       Y_{1}(p)=(h_{1}\ast X)(p) \textrm{ and } 
       Y_{2}(p)=(h_{2}\ast X)(p).
\end{equation*}
The corresponding spectral densities then are given by :
\begin{equation*}
       S_{Y_{1}}(\omega)=|H_{1}(\omega)|^2 S_X(\omega)
\end{equation*} 
and
\begin{equation*}
       S_{Y_2}(\omega)=16 |H_{1}(2\omega)|^2 S_X(\omega).
\end{equation*}
Assuming stationarity \cite{Rosenblatt1985} and after some Fourier domain
calculations, one gets that if $X$ has indeed an spectral
density as \eqref{eq:1} then the variances $\sigma_1$ and $\sigma_2$ of $Y_1$ and $Y_2$
respectively, are related by:
\begin{equation}\label{eq:2}
       \sigma_2 ^2 =2^{a+2} \sigma_1 ^2.
\end{equation}
Equation \eqref{eq:2} is the key to obtain estimates for $a$. In particular, this scale relation makes the estimation procedure easy to implement by wavelet filters \cite{Mall, Mey1992}. Furthermore, recalling that if $R_X$ is the correlation function of $X$ then $R_X$ is related by its Fourier Transform with $S_X$, i.e. $\widehat{R_X}=S_X$, we can prove a kind of reciprocal of this result:

%\begin{sat} 
\begin{theorem}Let $R_X (p)$ be a radial function and decreasing in $\|p\|$. If for some $a>0$ equation \ref{eq:2} holds for every $h_1 \in L^2(\mathbb{R}^2, S_X d\omega)$ then $\int\limits_{\mathbb{R}^2} |R_X(p)|dp=\infty$.  
\end{theorem}
%\end{sat}

The divergence of the integral of the correlation function is one of the usual forms to describe random fields with long range dependence. Thus, if an equation like \eqref{eq:2} holds for some fixed $a$ and any finite
energy filter then $X$ displays long range dependence, and moreover, with additional conditions equation \ref{eq:1} holds. In
practice, $a$ is unknown and one can only test with a finite
number of filters. However, based in these calculations, in
this work we propose that for a finite suitable parametrized
family of wavelet filters $h(s,p) , s=s_1,...,s_n$ and corresponding
estimates of $a(s)$ we can design a rather simple first warning
detection scheme which in a first stage heavily relies only on
linear, matrix and spectral numerical operations. 

\begin{figure}[ht]
\centering
\subfigure[Original image]{
\includegraphics[width=8.0cm,height=8.0cm]{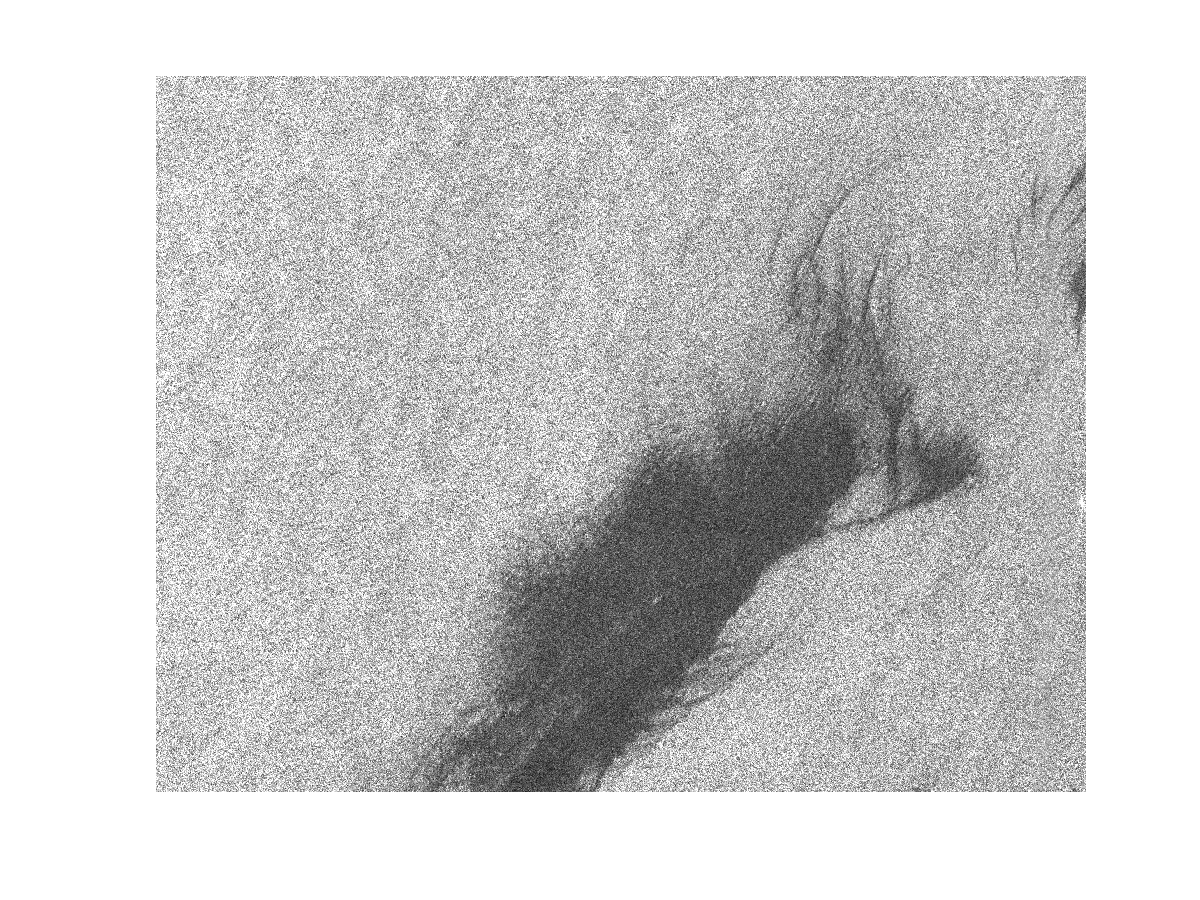} }
\subfigure[RGB map of estimated $\alpha$]{
\includegraphics[width=8.0cm,height=8.0cm]{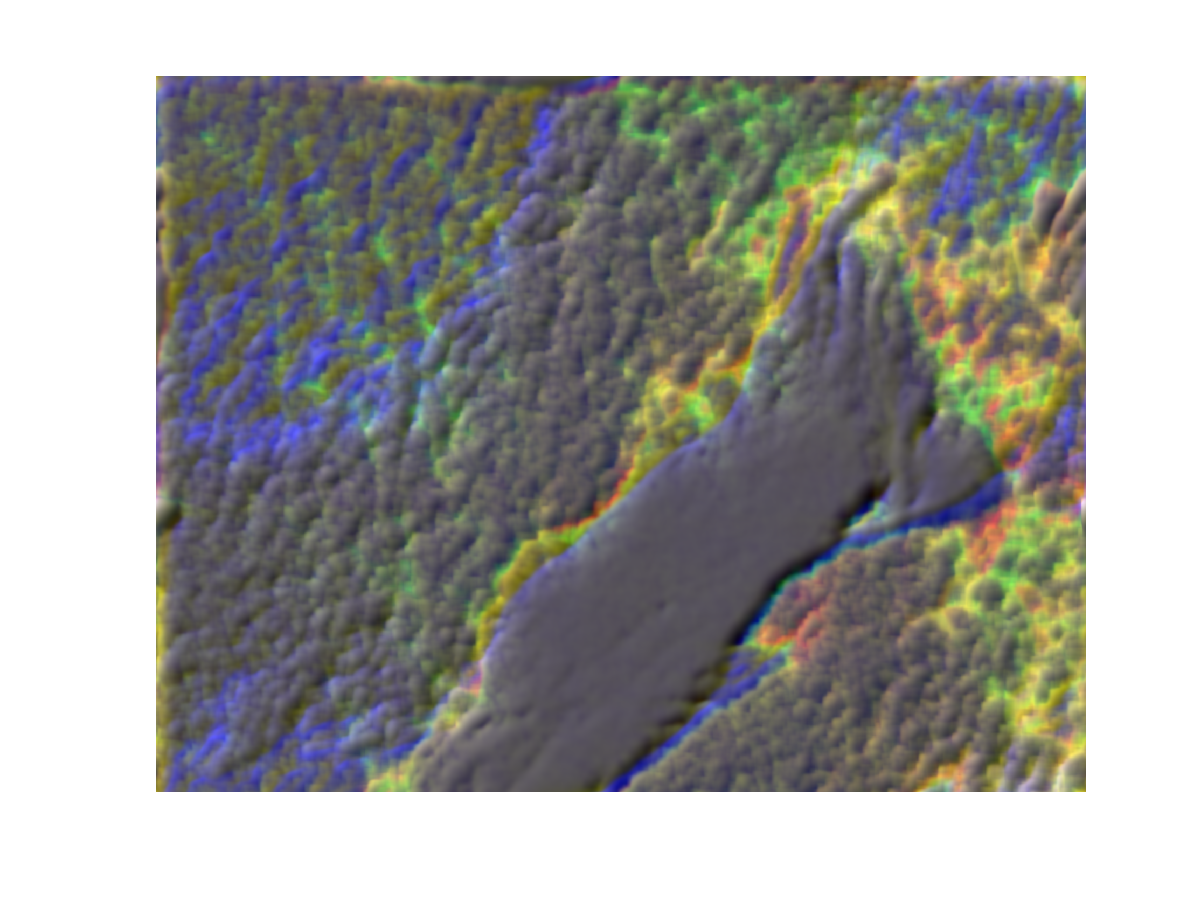} }
\subfigure[Neural Network Output stage]{
\includegraphics[width=8.0cm,height=8.0cm]{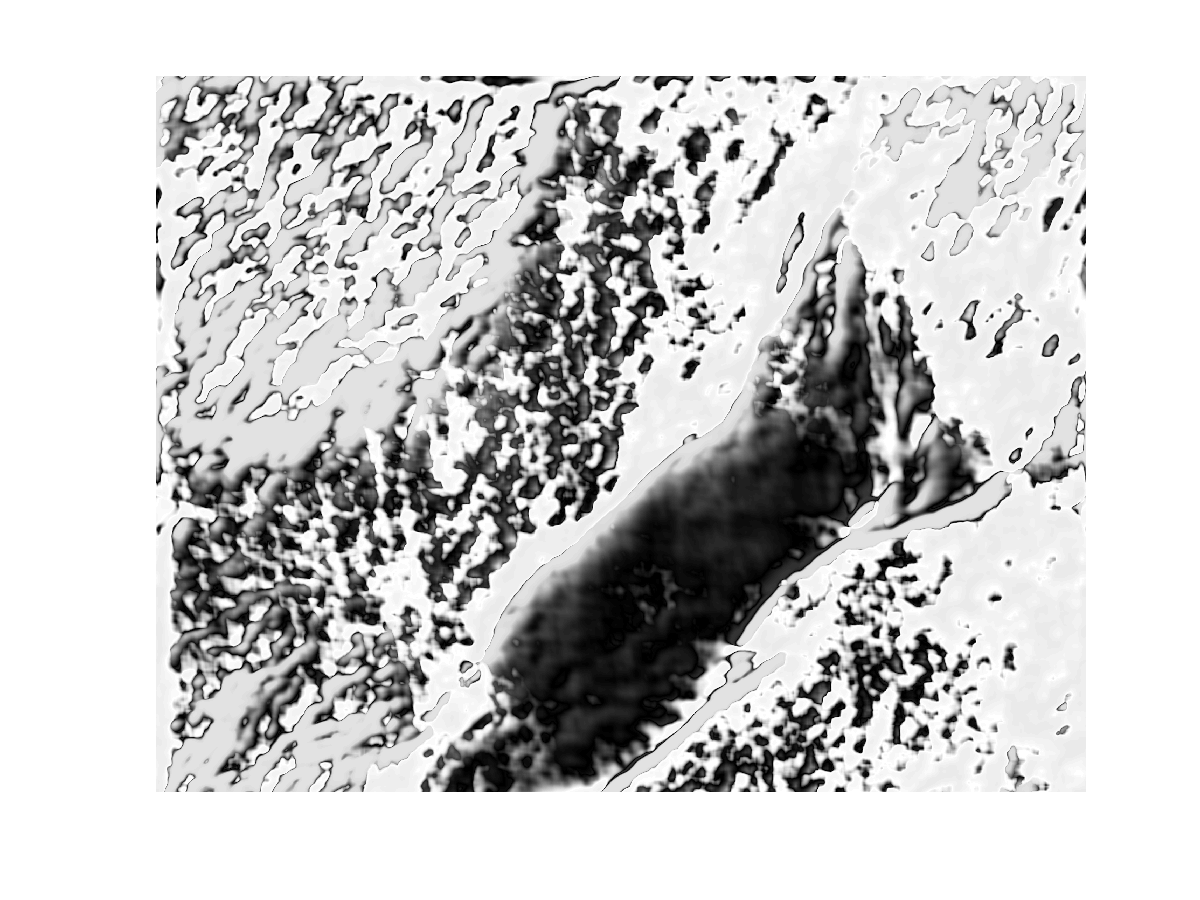}
}
\subfigure[
Hypotesis Test
Classification Stage. ~~~~~~~~~~~~~~~~~~~~~~
White/Positive {$=$} Oil Spill
]{
\includegraphics[width=8.0cm,height=8.0cm]{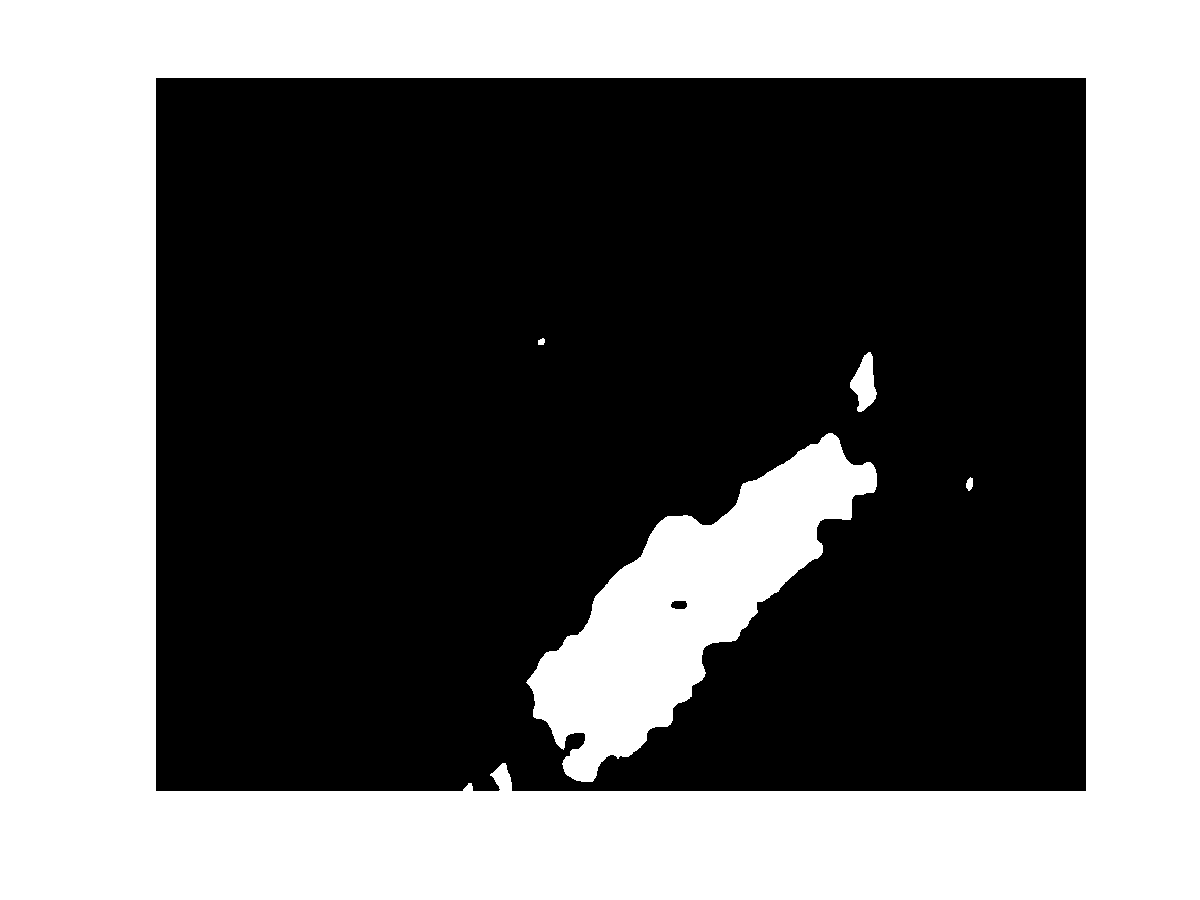}
}
\caption{Image analysis with Gaussian wavelet filters for Sentinel 1 Caspian Sea Oil Spill}
\label{fig:f1}
\end{figure}

Here we
present some practical examples with real images (see Figure \ref{fig:f1}). We note
that $a(s,p)$ defines in some way a transformation or map of
the original $X$. To illustrate the idea, if for certain fixed values
$s_1, s_2, s_3$ we generate an RGB auxiliary image assigning to
each color component a value linearly related to the
corresponding estimates of $a(s)$ for each filter. In the zones
of the image where these values are, ideally, equal the
related RGB intensities must be equal as well generating a
nearly dark grey flat area where $a(s)$ equals a constant.
Obviously if more filters $(n\geq 3)$ are involved more accurate
the following decision will become. The second stage
(classification and decision) takes the transformed image
$a(s,p)$ as an input into a combination of a neural network
and a classical statistical hypothesis \cite{DudaHartStork2000} testing previously
trained with several images samples and its respective
transforms $a(s,p)$. In this work we present some promising
experimental results with this technique.

\section*{Acknowledgements}
This work was funded by the Universidad de Buenos Aires, Grant. No. 20020170100266BA, CONICET and CONAE, under Project No. 5 of the \emph{Anuncio de Oportunidad para el desarrollo de aplicaciones y puesta apunto de metodolog\'{\i}as para el \'{a}rea oceanogr\'{a}fica utlizando im\'{a}genes SAR}, Buenos Aires, Argentina.


\begin{thebibliography}{1}
%\begin{thebibliography}{10}
%\bibitem{Hal50}
%P.~Halmos.
%\newblock {\em Measure Theory}.
%\newblock Van Nostrand, 1950.
%\bibitem{Gar15}
%A.~G. Garcia and M.~Mu\~{n}oz Bouzo.
%\newblock Sampling-related frames in finite u-invariant subspaces.
%\newblock {\em Appl. Comput. Harmon. Anal.}, 39,1, pp.173-184, 2015.
%\bibitem{Adams} Adams R.A., \textit{Sobolev Spaces}, %Academic Press, 1975.

\bibitem{Cohen} Cohen S., Istas J., \emph{Fractional Fields and Applications}, Springer, 2013.

\bibitem{DudaHartStork2000} Duda R., Hart P., and Stork D. \emph{Pattern Classification}, Wiley, 2nd. edition. 2000.

%\bibitem{Fal} Falconer K., \textit{Techniques in fractal %geometry}, Wiley, 1997.

%\bibitem{Flan} Flandrin P., ``Wavelet analysis and synthesis %of fractional Brownian motion'', \textit{IEEE Trans. Inf. %Theory}. IT 38(2), pp. 910-917, 1992.

%\bibitem{Gel} Gel'fand I.M. Vilenkin N. Ya. 
%\textit{Generalized Functions}. Vol. IV. Fizmatgiz, Moscow, %1961.(Russian).
%English trnsl. Academic Press, New York, 1964.

%\bibitem{Gikh} Gikhman I.I., Skorokhod A.V.\textit{Introduction to the theory of random processes}, Dover, 1996.

%\bibitem{Gra} Grafakos L. \textit{Classical Fourier Analysis}. Vol.I, GTM 249, Second Edition, Springer, 2008.

%\bibitem{Gra2} Grafakos L. \textit{Modern Fourier Analysis}. Vol.I, GTM 250, Second Edition, Springer, 2008.

%\bibitem{Lee} Lee A. J., ``Sampling theorems for non stationary processes'', \textit{Trans. of the A.M.S.}  V. 242, 1978, pp.225-241.

%\bibitem{Kah} Kahane J.P. \textit{Some random series of functions}, Cambridge Studies in Advance Mathematics No. 5, Cambridge University Press, (2nd edition) 1994.

\bibitem{Mall}Mallat S. \textit{A Wavelet tour of signal processing}, Academic Press, 2008.

\bibitem{MarghanyHashim2007} Marghany M., Hashim M., \emph{Fractal dimension algorithm
for detection Oil Spills using RADARSAT 1 SAR}, in
Computational Science and its applications ICCSA 2007.
Lecture Notes in Computer Science, Vol. 4705. Springer,
2007.

\bibitem{MarghanyHashim2011} Marghany M., Hashim M. \emph{Discrimination between oil spill
and look alike using Fractal dimension algorithm form
RADARSAT 1 SAR and AIRSAR/POLSAR data}, International
Journal of Physical Sciences , 6(7), pp. 1711-1719, 2011.


%\bibitem{Mas} Masry E. ``The wavelet transform of stochastic processes with stationary increments and its applications to fractional Brownian motion''. \textit{IEEE Trans. on Inf. Theory}. IT 34(1), pp. 260-264, 1993.

%\bibitem{yo2} Medina J.M. Cernuschi-Fr\'{\i}as B. ``On the a.s. convergence of certain random series to a fractional random field in $\mathcal{D}^{\prime}(\mathbb{R}^d)$''. \textit{Statistics and Probability Letters}, 74(2005), pp. 39-49.

%\bibitem{yo2} Medina J.M., Dobarro F.R., Cernuschi-Fr\'{\i}as B. ``Convergence of $p$-Stable Random Fractional Wavelet Series and Some of its Properties''. (submitted)

%\bibitem{art2} Meyer Y., Sellan F., Taqqu M.S., ``Wavelets, generalized white noise and fractional integration: The synthesis of Fractional Brownian Motion'' \textit{The Journal Of Fourier Analysis and  Applications}, Vol 5, Issue 5,1999.%

\bibitem{Mey1992} Meyer Y. \textit{Wavelets and operators}. Cambridge Studies in Advance Mathematics No. 37, Cambridge University Press, 1992.

%\bibitem{Pip} Pipiras V., Taqqu M.S., Abry P., ``Can continuous time stable processes have discrete linear representations?'',  \textit{Statistics and Probability Letters}, 64(2003), pp. 147-157.

\bibitem{Rosenblatt1985} Rosenblatt M., Stationary Sequences and Random Fields, Birkhäuser, 1985.


%\bibitem{Sam}  Samorodnitsky G. and Taqqu M.S., \textit{Stable Non-Gaussian Random Processes} Chapman and Hall/CRC, 1994.

%\bibitem{Stein} Stein E.M., \textit{Singular Integrals and Differentiability Properties of functions}, Princeton Univ. Press (1970)

%\bibitem{Taqq} Samorodnitsky G., Taqqu M.S., \emph{Stable Non-Gaussian Random Processes}, Chapman and Hall-CRC, 1994.

%\bibitem{Tafti1} Tafti P.D., Unser M., ``Fractional Brownian vector fields'', \textit{SIAM Multiscale Model. Simul.} 8(5), pp. 1645-1670, 2010.

%\bibitem{Unser} Unser M., Tafti P., \emph{An Introduction to Sparse Stochastic Processes}, Cambridge, 2014.

%\bibitem{Van} Van de Ville D., Tafti P.D., Unser M., ``Invariance, Laplacian-Like Wavelet Bases'', \textit{IEEE Trans. Image Process.} 18(4), pp. 689-702, 2009.

\end{thebibliography}
\end{document}